\PassOptionsToPackage{bookmarks={false},hidelinks}{hyperref}
\documentclass[conference,10pt]{IEEEtran}

\usepackage{cite}
\usepackage{amsmath,amsfonts,amsthm}
\usepackage[ruled]{algorithm}
\usepackage[noend]{algpseudocode}
\usepackage[normalem]{ulem}
\usepackage{graphicx}
\usepackage{textcomp}
\usepackage{xcolor}
\usepackage{color}
\usepackage{etoolbox}
\usepackage{subcaption}
\usepackage{multirow}
\usepackage[misc,geometry]{ifsym}

\usepackage{amsmath,amsfonts,bm}

\def\eqref#1{equation~\ref{#1}}

\def\1{\bm{1}}

\def\mX{{\bm{X}}}
\def\mY{{\bm{Y}}}

\usepackage{hyperref}
\usepackage{url}
\usepackage[english]{babel}

\usepackage{booktabs}
\def\BibTeX{{\rm B\kern-.05em{\sc i\kern-.025em b}\kern-.08em
    T\kern-.1667em\lower.7ex\hbox{E}\kern-.125emX}}

\newbool{inccomment}
\setbool{inccomment}{true}

\begin{document}
\title{CSQ: Growing Mixed-Precision Quantization Scheme with Bi-level Continuous Sparsification}

\author{
Lirui Xiao$^{*,1}$, Huanrui Yang$^{*,2}$, Zhen Dong$^2$, Kurt Keutzer$^2$, Li Du\textsuperscript{\Letter}$^{,1}$, Shanghang Zhang\textsuperscript{\Letter}$^{,3}$ \\
$^1$Nanjing University, \
$^2$University of California, Berkeley, \
$^3$Peking University\\
{\tt\small \{lirxiao,ldu\}@nju.edu.cn \{huanrui,zhendong,keutzer\}@berkeley.edu}\\ 
{\tt\small shanghang@pku.edu.cn }
}

\maketitle
\newcommand\blfootnote[1]{%
\begingroup
\renewcommand\thefootnote{}\footnote{#1}%
\addtocounter{footnote}{-1}%
\endgroup
}

\blfootnote{* Equal contribution.}
\blfootnote{\textsuperscript{\Letter} Corresponding Author.}

\begin{abstract}
Mixed-precision quantization has been widely applied on deep neural networks (DNNs) as it leads to significantly better efficiency-accuracy tradeoffs compared to uniform quantization. Meanwhile, determining the exact precision of each layer remains challenging. Previous attempts on bit-level regularization and pruning-based dynamic precision adjustment during training suffer from noisy gradients and unstable convergence.
In this work, we propose Continuous Sparsification Quantization (CSQ), a bit-level training method to search for mixed-precision quantization schemes with improved stability. CSQ stabilizes the bit-level mixed-precision training process with a bi-level gradual continuous sparsification on both the bit values of the quantized weights and the bit selection in determining the quantization precision of each layer.
The continuous sparsification scheme enables fully-differentiable training without gradient approximation while achieving an exact quantized model in the end.
A budget-aware regularization of total model size enables the dynamic growth and pruning of each layer's precision towards a mixed-precision quantization scheme of the desired size.
Extensive experiments show CSQ achieves better efficiency-accuracy tradeoff than previous methods on multiple models and datasets.

\end{abstract}

\section{Introduction}
\label{sec:intro}

With the wide application of deep neural networks (DNNs) in mobile and edge applications~\cite{sandler2018mobilenetv2,hu2018squeeze}, improving the efficiency of DNNs has been extensively researched. 
Quantization, which converts the weight and activation of the DNN model from high-precision floating point values to low-precision fixed-point representations, has been widely used to improve DNN efficiency~\cite{zhou2016dorefa,choi2018pact,zhang2018lq}. 
Besides largely reducing the number of bits to store the model, the fixed-point representation achieved by linear quantization also enables the use of fixed-point arithmetic units, which largely reduces the area and energy cost, and leads to significant speedup compared to the floating-point counterparts~\cite{horowitz20141}.

Nevertheless, the quantization process introduces perturbation to the optimal weight value, which hinders the performance of quantized DNNs. To improve quantized model performance, previous research identifies that not all layers in a DNN are equally sensitive to quantization~\cite{dong2019hawq,yang2021bsq}, which leads to the idea of mixed-precision quantization: Sensitive layers are allowed to keep higher precision, while less sensitive layers are quantized to lower precision, therefore reaching a better model size-performance tradeoff.


Due to the large and discrete design space, the difficulty of mixed-precision quantization lies in determining the exact precision of each layer. Previous work tackles the precision assignment problem via reinforcement learning-based search~\cite{wang2019haq} or utilizes higher-order sensitivity statistics computed on the pretrained model~\cite{dong2019hawq,dong2020hawq}. However, the search-based method is costly to run, and the statistics in the pretrained model do not capture the potential sensitivity changes during the model training process. Dynamically achieving a mixed-precision quantization scheme during training is also attempted through the lens of bit-level structural sparsity~\cite{yang2021bsq}, yet the bit-level training process and the periodic precision adjustment in training both lead to an unstable convergence~\cite{yang2021bsq}.

In this work, we aim to improve the stability of bit-level training and precision adjustment to achieve a better convergence toward a mixed-precision quantized DNN. We locate two main factors of the instability: 1) the binary selection of bit value, and 2) the binary selection of using a certain bit or not in determining the precision of each layer. Previous methods approximate the gradient of these discrete selections via straight-through estimator (STE)~\cite{bengio2013estimatingSTE}, which can be noisy and hinders convergence. 
Instead, this work proposes Continuous Sparsification Quantization (CSQ). CSQ utilizes the idea of continuous sparsification~\cite{savarese2020winning,yuan2020growing} to relax both levels of discrete selection with a series of smooth parameterized gate functions. The smoothness enables fully differentiable training of the bit-level model without gradient approximation, while proper scheduling of the gate function parameter enables the model to converge to an exact quantized form without additional rounding. We further integrate budget-aware regularization on the bit selection into the pipeline, in order to induce a mixed-precision quantization scheme under the budget constraint through an end-to-end differentiable training process.
To the best of our knowledge, CSQ is the first to make the following contributions:
\begin{itemize}
    \item Utilize continuous sparsification technique to improve bit-level training of quantized DNN;
    \item Relax precision adjustment in the search of mixed-precision quantization scheme into smooth gate functions;
    \item Combine the bi-level continuous sparsification into effectively inducing high-performance mixed-precision DNNs. 
\end{itemize}

The effectiveness of CSQ is well supported by extensive empirical results. For instance, on the CIFAR-10 dataset, CSQ achieves 1.5$\times$ further compression over BSQ~\cite{yang2021bsq} for ResNet-20 model under similar accuracy, and achieves 16$\times$ lossless compression for the VGG19BN model. On ImageNet, CSQ achieves a lossless 10.7$\times$ compression for the ResNet-18 model, which is 0.43\% higher top-1 accuracy than the same-sized LQ-Net~\cite{zhang2018lq}. For ResNet-50, CSQ leads to a 17\% further compression than BSQ at the same accuracy.
\section{Related work}
\label{sec:back}

\subsection{Mixed-precision quantization}

The key research question of mixed-precision quantization has been how to design a set of bit schemes that achieve the best performance-size tradeoff. Early attempts use manual design heuristics such as keeping the first and last layer at a higher precision~\cite{rastegari2016xnor,li2016ternary}. 
Searching-based methods like HAQ~\cite{wang2019haq} utilize reinforcement learning to determine the quantization scheme, yet the search cost is often high, especially for deeper models with an exponentially large search space.
Another line of methods directly measures the sensitivity of each layer with metrics like Hessian eigenvalue~\cite{dong2019hawq} or Hessian trace~\cite{dong2020hawq}. 
However, such methods only incorporate the sensitivity of the pretrained full-precision model, without considering the change of sensitivity when the weights are being quantized or being updated in the quantization-aware training process. 
Bit-level sparsity quantization (BSQ)~\cite{yang2021bsq} makes the first attempt to simultaneously induce mixed-precision quantization scheme and train the quantized DNN model within a single round of training. BSQ considers each bit of the quantized model as independent trainable variables, and achieves mixed-precision quantization scheme by inducing bit-level structural sparsity. The bit-level representation of layer weight $W$ can be formulated as:
\begin{equation}
\label{equ:BSQ}
    W = \frac{s}{2^n-1}\text{Round}\left[\sum_{b=0}^{n-1}\left(W_p^{(b)}-W_n^{(b)}\right)2^b \right],
\end{equation}
where $s$ is the scaling factor, $W_p^{(b)}$ and $W_n^{(b)}$ are the $b$-th bit of the positive and negative values in $W$ respectively, and $n$ is the quantization precision of the layer.
Though BSQ leads to good empirical results, the rounding in the bit-level representation requires straight-through gradient estimation~\cite{bengio2013estimatingSTE} on the bit variables, which can be inaccurate. Also, the hard precision adjustment performed via bit pruning during training hinders the convergence stability. In this work, we relax both bit-level training and precision adjustment with continuous sparsification, leading to improved stability and performance over BSQ.

\subsection{Sparse optimization and continuous sparsification}

The difficulty of optimizing discrete values isn't only faced by quantization research. Research on DNN pruning also needs to accommodate the binary mask of selecting a weight element/filter or not into the model training process. Minimizing the $\ell_0$ regularization, which is the sum of the binary weight selection mask, has been identified as a straightforward and unbiased method to induce sparse neural network~\cite{louizos2017learning}, but is difficult due to the discrete nature of the mask. Therefore attempts have been made to relax the binary constraint on the mask to enable gradient-based training.

Louizos et al. first propose to consider the binary mask as stochastic gates, whose distribution can then be relaxed into ``Hard concrete distribution''~\cite{louizos2017learning} or ``Scale mixture of Gaussian''~\cite{louizos2017bayesian} with learnable parameters. However, the gradient estimation required in stochastic optimization leads to high variance and performs poorly on larger models.
Continuation methods, on the other hand, approximate the binary constraint by relaxing it with a smooth gating function, while gradually making closer approximations to the binary gate as training progresses. For instance, continuous sparsification~\cite{savarese2020winning,yuan2020growing} relax the binary gate $I(x\geq 0)$ as a Sigmoid function with temperature as formulated in Equation~(\ref{equ:sigmoid}).
\begin{equation}
\label{equ:sigmoid}
    I(x\geq 0) \sim f_\beta(x) = \sigma(\beta x) = \frac{1}{1+e^{-\beta x}}.
\end{equation}
Temperature $\beta$ controls the smoothness of the relaxed gate, which grows exponentially with the training epochs. A smaller $\beta$ is used at early epochs to enable a smooth optimization. A larger $\beta$ is used in later epochs to better approximate the discrete binary gate. 
As continuous sparsification is mainly explored under the DNN pruning setting, this work serves as the first attempt to apply bi-level continuous sparsification on both bit-level training and bit selection to induce a mixed-precision quantized DNN model.

\section{Method}

This section introduces our method of growing mixed-precision quantized models. Section~\ref{ssec:CS} formulates the bi-level continuous sparsification of the bit-level representation for smooth optimization. Section~\ref{ssec:Budget} describes the budget-aware model size regularization that controls the growth and prune of layer precision.
The overall algorithm of CSQ is provided in Section~\ref{ssec:overall}.

\subsection{Bi-level continuous sparsification of quantized DNN model}
\label{ssec:CS}

\begin{figure}[tb]
\centering
\includegraphics[width=\linewidth]{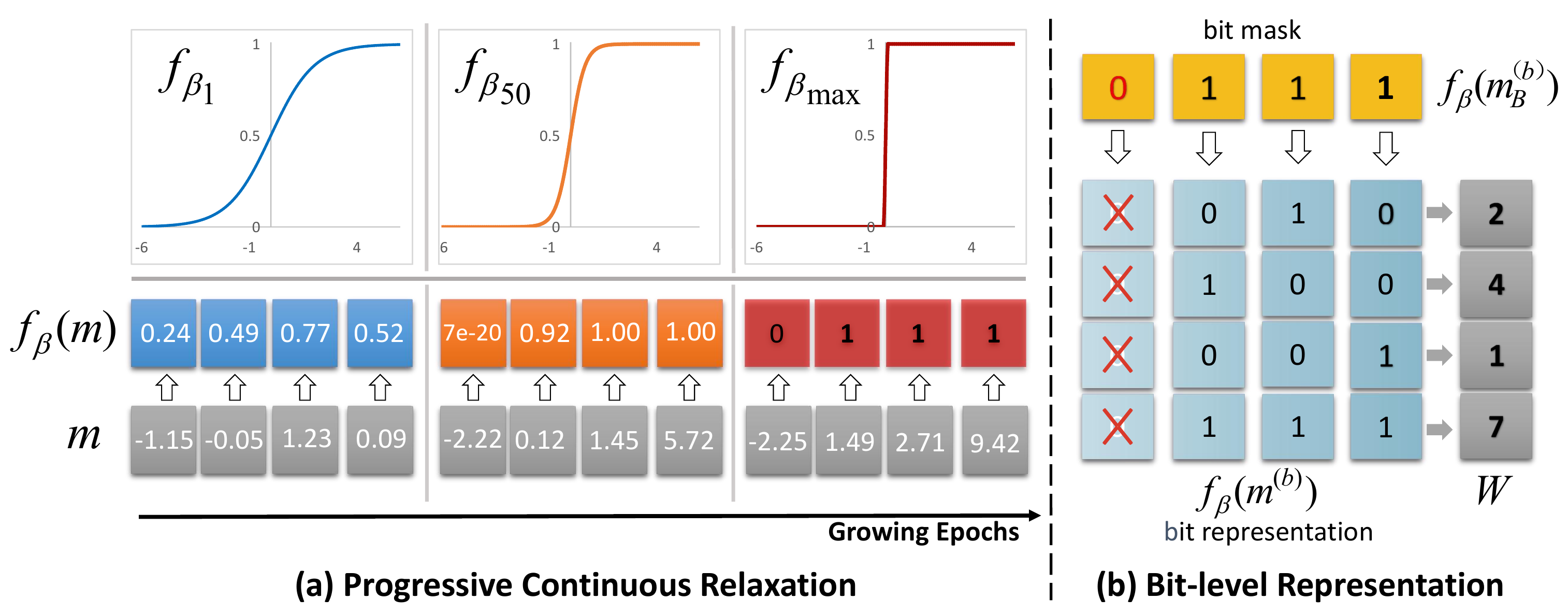}
\caption{Illustrating the bi-level continuous sparsification used in CSQ. (a) The change of temperature sigmoid gating function throughout the training process. (b) Representing quantized model with bit mask and bit representation.}
\label{fig:pipeline}
\vspace{-10pt}
\end{figure}

To represent a quantized DNN model, we need: 1) the quantization precision of each layer, and 2) the quantized value of each weight element. Both properties take discrete values, which prevent direct gradient-based updates. This work aims to relax the discrete optimization of these properties with a continuous differentiable function, enabling a smooth and differentiable optimization.

Specifically, inspired by previous attempts on continuous sparsification for DNN pruning~\cite{yuan2020growing}, we utilize the temperature Sigmoid function $f_\beta(\cdot)$ defined in Equation~(\ref{equ:sigmoid}) to relax the binary representation, where $f_\beta$ can directly replace the bit-level weight $W_p^{(b)}$ and $W_n^{(b)}$ defined in Equation~(\ref{equ:BSQ}). For example, consider a layer with weight tensor $W$ under a linear symmetric $n$-bit quantization, the quantized weight can be relaxed as 
\begin{equation}
\label{equ:bit}
    W = \frac{s}{2^n-1}\sum_{b=0}^{n-1}\left[\left(f_\beta\left(m_p^{(b)}\right)-f_\beta\left(m_n^{(b)}\right)\right)2^b \right],
\end{equation}
where $f_\beta(m_p^{(b)})$ and $f_\beta(m_n^{(b)})$ are the relaxed bit-level representation of the positive and negative values in $W$ respectively, with $m_p^{(b)}$ and $m_n^{(b)}$ taking any real values. 

In training, we consider $s, m_p^{(b)}, m_n^{(b)}$ instead of $W$ as trainable variables. We exponentially increase the value of $\beta$ with the number of epochs. In this way, the trainable variables can be optimized smoothly in the early training stage, while gradually converging to an exact quantized model as $f_\beta(\cdot)$ converges to a unit step function with $\beta \rightarrow +\infty$, as illustrated in Figure~\ref{fig:pipeline}(a). Since no rounding is applied, there's no approximation of gradient required.

With the quantized model representation in hand, the precision of each layer can be controlled as selecting the number of bits to be used. This can be formulated as having a binary bit mask $q_B\in \{0,1\}^n$ in each layer as
\begin{equation}
\label{equ:bit_mask}
    W = \frac{s}{2^n-1}\sum_{b=0}^{n-1}\left[\left(f_\beta\left(m_p^{(b)}\right)-f_\beta\left(m_n^{(b)}\right)\right)2^b q_B^{(b)} \right],
\end{equation}
where $q_B^{(b)}=1$ if the bit is selected and 0 otherwise. Thus the precision of the layer can be computed as $\sum_b q_B^{(b)}$. 

Since $q_B$ is binary, it can also be relaxed with continuous sparsification. The resulting quantized model is formulated in Equation~(\ref{equ:bit_mask_CS}), and demonstrated in Figure~\ref{fig:pipeline}(b).
\begin{equation}
\label{equ:bit_mask_CS}
    W = \frac{s}{2^n-1}\sum_{b=0}^{n-1}\left[\left(f_\beta\left(m_p^{(b)}\right)-f_\beta\left(m_n^{(b)}\right)\right)2^b f_{\beta}\left(m_B^{(b)}\right) \right].
\end{equation}
Here we can use the same temperature scheduling for both bit masks and bit representations of each layer.

\subsection{Budget-aware growing of mixed-precision quantization scheme}
\label{ssec:Budget}
Now that we have relaxed both bit representation and bit selection of a quantized DNN in Equation~(\ref{equ:bit_mask_CS}), the next step is to adjust the precision of each layer so that it can achieve a mixed-precision quantization scheme within the available budget. This can be achieved with an $\ell_1$ regularization over the bit-mask of each layer, as
\begin{equation}
    R(m_B) =  \sum_b f_{\beta}\left(m_B^{(b)}\right).
\end{equation}

With the regularization, the final training objective is
\begin{equation}
\label{equ:obj}
    \min_{s,m_p,m_n,m_B} \mathcal{L}(W) + \lambda \Delta_S \sum_{\text{Layer}} R(m_B), 
\end{equation}
where $\mathcal{L}(\cdot)$ is the original loss of DNN training, $W$ is the model weight parameterized as Equation~(\ref{equ:bit_mask_CS}), $\lambda$ is the base regularization strength, and $\Delta_S$ is the budget-aware scaling factor. $\Delta_S$ is added to encourage more pruning when that current model size is significantly large than the budget, less pruning if the current model size is close to the budget, and growing (have a negative regularization) when the current model size is smaller than the budget. Therefore we define $\Delta_S$ as the average quantization precision of all elements in the current model minus the targeted average precision of the budget. We count the precision of each layer during the training process based on the value of $m_B$, where the precision is determined as $\sum_b \left[ m_B^{(b)}\geq 0 \right]$.
The training objective is end-to-end differentiable with respect to the scaling factor $s$, bit representation $m_p, m_n$, and bit mask $m_B$ of each layer, without the need to apply gradient approximation via straight-through estimation.

\subsection{Overall training algorithm}
\label{ssec:overall}

Putting everything together, performing stochastic gradient descent with the derived objective in Equation~(\ref{equ:obj}) leads to the CSQ algorithm, as illustrated in detail in Algorithm~1. 
Bit representations and bit masks are trained simultaneously, leading to a joint optimization for both model weight values and quantization precision. The sigmoid temperatures $\beta$ is scheduled to grow exponentially with the training epochs, gradually converting the model with smooth gating functions into an exactly quantized model, where $f_\beta$ converges to a unit-step Sign function $I(m\geq0)$ at $\beta_{\max}$.

For complicated tasks like ImageNet, we can further boost the final model performance by applying an additional finetuning of the achieved mixed-precision quantized model. During the finetuning process, we fix the quantization scheme of each layer, while only tuning the bit representation $s, m_p, m_n$ of the selected bits in each layer. We rewind the temperature $\beta$ for the bit representation back to $1$, and redo the exponential temperature scheduling with the number of finetuning epochs. This process gives the bit representations adequate flexibility to further improve the model performance under the mixed-precision quantization scheme found by CSQ.

\renewcommand{\algorithmicrequire}{\textbf{Input:}}
\renewcommand{\algorithmicensure}{\textbf{Output:}}
\begin{algorithm}
\small
        \caption{Bi-level continuous sparsification}
        \begin{algorithmic}
            \Require Data $\mX = (x_i)^n_{i=1}$, labels $\mY = (y_i)^n_{i=1}$
            \Ensure mixed-precision model $G$
            \State Initialize: $s, m_p, m_n, m_B$ in $G$
            \State Initialize temperature: $\beta_0 = 1$, set $\beta_{\max} = 200$
            \State \textit{\color[rgb]{0,0.4,0}\# CSQ Training}
            \For {$epoch=0,\ldots,T$} 
            \State Temperature scheduling $\beta = \beta_0 \beta_{\max}^{epoch/T}$
                    \For {$i=i,\ldots,n$}
                        \State Sample mini-batch $x_i,y_i$ from $\mX,\mY$
                        \State Compute model weight $W$ using Eq.~(\ref{equ:bit_mask_CS})
                        \State Update trainable parameters with Eq.~(\ref{equ:obj})
                    \EndFor
            \EndFor
        \State \textit{\color[rgb]{0,0.4,0}\# Mixed-precision finetuning (optional)}
        \State Fix bit selection $q_B^{(b)}=I(q_B^{(b)}\geq0)$
        \State Temperature rewind $\beta=\beta_0$
        \For {$epoch=0,\ldots,T'$} 
            \State Temperature scheduling $\beta = \beta_0 \beta_{\max}^{epoch/T'}$
                    \For {$i=i,\ldots,n$}
                        \State Sample mini-batch $x_i,y_i$ from $\mX,\mY$
                        \State Compute model weight $W$ using Eq.~(\ref{equ:bit_mask})
                        \State Update $s, m_p, m_n$ with $\mathcal{L}(W)$
                    \EndFor
            \EndFor
        \State return $G$
        \end{algorithmic}
\end{algorithm}
\vspace{-10pt}
\section{Evaluation}
This section summarizes the empirical results of CSQ. We evaluate CSQ using ResNet-20~\cite{he2016deep} and VGG19BN~\cite{simonyan2014very} on CIFAR-10~\cite{krizhevsky2009learning}, and using ResNet-18 and ResNet-50~\cite{he2016deep} on ImageNet~\cite{deng2009imagenet}. We compare the results of our method with existing uniform~\cite{zhang2018lq,zhou2016dorefa,choi2018pact} and mix-precision~\cite{dong2019hawq,yao2021hawq,wang2019haq,yang2021bsq} quantization methods. Ablation studies are also provided.

\subsection{Experimental Setup}
We use the same set of hyperparameters for experiments conducted on the same model. All models are trained with SGD with an initial learning rate of 0.1 and a cosine annealing schedule. We use a linear learning rate warm-up for the first 5 epochs for ImageNet experiments. Weight decay is set to $5e-4$ for CIFAR-10 and $1e-4$ for ImageNet. Momentum is set to 0.9. All models are trained from scratch. On CIFAR-10, ResNet-20 model is trained with CSQ for 600 epochs, and VGG for 300 epochs, without finetuning. ImageNet models are trained with 200 CSQ epochs plus 100 epochs of finetuning after finalizing the quantization scheme, as introduced in Algorithm~1. These training epochs are comparable with the total training epochs (pretraining + finetuning) used in previous methods like BSQ~\cite{yang2021bsq} and HAWQ~\cite{dong2019hawq}.

For the CSQ training process, we set the shape of the bit representation and bit mask to uniform 8-bit in each layer, as in most cases 8-bit is adequate for a lossless quantization. 
Since CSQ does not control activation quantization, we quantize the activation uniformly throughout the training process, whose precision is reported in the ``A-Bits'' column in the tables.
We set the base regularization strength $\lambda$ as 0.01 for training all models. The maximum temperature of the soft gate $f_\beta$ function for both bit representation weight and bit mask is set as 200, which will be reached in the last epoch. At the last epoch $f_\beta$ will turn into a steep step function, where all of its output should be either 0 or 1. Additionally, to ensure an exactly quantized model at the end of the training, we set all gate functions to the unit-step function before the final validation. 

\subsection{Experimental Results}
This section compares our results with previous quantization methods.
In all tables ``FP'' refers to the full-precision model; ``MP'' denotes mixed-precision weight quantization; and ``T'' denotes the target precision of CSQ. Weight compression ratio ``Comp'' is computed with respect to the full-precision model.

{\noindent \textbf{CIFAR-10 results.}} For ResNet-20 on CIFAR-10, we compare CSQ with various previous methods. As shown in Table~\ref{table1}, CSQ outperforms previous methods under all activation precision. Notably, the CSQ-T2 model with 3-bit activation enables an 1.5$\times$ further compression vs. BSQ~\cite{yang2021bsq} under the same accuracy.
Similarly, CSQ also demonstrated superior performance on the VGG19BN model, as shown in Table~\ref{table2}. Under full-precision activation CSQ enables a nearly lossless 16$\times$ compression, largely pushing the frontier of previous methods. Furthermore, CSQ method even surpasses non-linear quantizers, which are generally powerful but unfriendly for implementation. Comparing to non-linear quantization methods LQ-Nets~\cite{zhang2018lq} and~\cite{gennari2020finding}, CSQ achieves both higher accuracy and higher compression ratio up to 1.8$\times$.

\begin{table}[tb]
\small
\centering
\caption{Quantization results of ResNet-20 models on the CIFAR-10 dataset.}
\label{table1}
\begin{tabular}{c | c c c c}
\toprule
A-Bits     & Method   & W-Bits    & Comp($\times$)  & Acc(\%)  
\\ 
\midrule
\multirow{3}{*}{32}  & FP & 32 & 1.00 & 92.62       
                         \\ 
                         & LQ-Nets\cite{zhang2018lq} & 3 & 10.67 & 92.00
                         \\
                         & BSQ\cite{yang2021bsq} & MP & 19.24 & 91.87
                         \\
                         & {\bfseries CSQ T1} & \textbf{MP} &\textbf{26.67}   &\textbf{91.70}
                         \\
                         & {\bfseries CSQ T2} & \textbf{MP} &\textbf{16.00}   &\textbf{92.68}
                         \\
                \hline
\multirow{3}{*}{3}   & LQ-Nets  &  3  & 10.67  &   91.60          
                         \\ 
                         & PACT\cite{choi2018pact}  & 3 & 10.67 & 91.10
                         \\
                         & DoReFa\cite{zhou2016dorefa} & 3 & 10.67 & 89.90
                         \\
                         & BSQ & MP & 11.04 & 92.16
                         \\
                         & {\bfseries CSQ T2} & \textbf{MP} &\textbf{16.93}   &\textbf{92.14}
                         \\
                         & {\bfseries CSQ T3} & \textbf{MP} &\textbf{10.49}   &\textbf{92.42}
                         \\
                \hline
 \multirow{3}{*}{2}  & LQ-Nets  & 2  &   16.00  & 90.20
                         \\
                         & PACT    & 2  &  16.00   & 89.70
                         \\
                         & DoReFa   & 2  &  16.00   & 88.20
                         \\
                         & BSQ      & MP & 18.85    & 90.19
                         \\
                         & {\bfseries CSQ T1} & \textbf{MP} &\textbf{22.86}   &\textbf{90.08}
                         \\
                         & {\bfseries CSQ T2} & \textbf{MP} &\textbf{16.41}   &\textbf{90.33}
                         \\
                \bottomrule
\end{tabular}
\end{table}

\begin{table}[tb]
\small
\centering
\caption{Quantization results of VGG19BN models on the CIFAR-10 dataset.}
\label{table2}
\begin{tabular}{c | c c c c}
\toprule
A-Bits.     & Method   & W-Bits.    & Comp($\times$)  & Acc(\%)    \\ 
\midrule
\multirow{3}{*}{32}  
& FP & 32 & 1.00 & 94.22 \\ 
& LQ-Nets~\cite{zhang2018lq}    &  3   &  10.67  &  93.80  \\
& {\bfseries CSQ T2} & \textbf{MP} &\textbf{16.00}   &\textbf{94.10} \\
\hline
    & ZeroQ~\cite{cai2020zeroq}     &  4  &   8.00   &   92.69    \\
8    & ZAQ~\cite{Liu_2021_CVPR}     &  4  &   8.00   &   93.06    \\
& {\bfseries CSQ T3} & \textbf{MP} &\textbf{10.67}   &\textbf{93.90}  \\
\hline
&  QUANOS~\cite{panda2020quanos}   &  MP   &  7.11  & 90.70   \\
4    
& {\bfseries CSQ T3} & \textbf{MP} &\textbf{10.67}   &\textbf{93.62}  \\
\hline
\multirow{3}{*}{3}   
& LQ-Nets~\cite{zhang2018lq}  &  3  &  10.67    &  93.80  \\ 
& Non-Linear~\cite{gennari2020finding}    &  3   &  9.14  &  93.40  \\
& {\bfseries CSQ T2} & \textbf{MP} &\textbf{16.00} &\textbf{93.58}    \\
\bottomrule
\end{tabular}
\vspace{-10pt}
\end{table}

{\noindent \textbf{ImageNet results.}} To evaluate the scalability of CSQ, we perform experiments on the large-scale ImageNet dataset on deeper models. Table~\ref{table3} shows the results of ResNet18 and ResNet-50 obtained by different quantization methods~\cite{zhou2016dorefa, choi2018pact, zhang2018lq, wang2019haq,yao2021hawq,yang2021bsq}, among which CSQ consistently shows strong performance. For ResNet-18, the model obtained by CSQ-T3 with an average of 3-bit weight precision and 8-bit activation precision achieves almost the same accuracy as the full-precision baseline. CSQ-T2, with 4-bit activation, achieves a higher compression rate (15.23$\times$) with a tiny accuracy drop. This result significantly outperforms the W4A4 uniformly quantized model reported by HAWQ-V3~\cite{yao2021hawq} with 1.9$\times$ further compression. For ResNet-50, CSQ also achieves better efficiency-accuracy tradeoff, beating strong mixed-precision quantization baseline HAQ~\cite{wang2019haq} and BSQ~\cite{yang2021bsq}.

\begin{table}[tb]
\small
\centering
\caption{ Quantization results of ResNet-18 and ResNet-50 models on the ImageNet dataset.}
\label{table3}
\resizebox{\linewidth}{!}{
\begingroup
    \setlength{\tabcolsep}{2pt}
\begin{tabular}{c|ccc|ccc} 
\toprule
\multicolumn{1}{c}{} & \multicolumn{3}{c}{ResNet-18} & \multicolumn{3}{c}{ResNet-50} \\ 
\hline
Method & W-Bits & Comp($\times$) & Acc(\%) & W-Bits & Comp($\times$) & Acc(\%)    \\ 
\hline
FP & 32 & 1.00 & \textcolor[rgb]{0.149,0.149,0.149}{69.76} & 32 & 1.00 & 76.13 \\ 
\hline
DoReFa\cite{zhou2016dorefa}    & 5     & 6.40      & 68.4        & 3     & 10.67     & 69.90     \\
PACT\cite{choi2018pact}      & 4     & 8.00      & 69.2        & 3     & 10.67     & 75.30     \\
LQ-Nets\cite{zhang2018lq}   & 3     & 10.67     & 69.30       & 3     & 10.67     & 74.20     \\
HAWQ-V3\cite{yao2021hawq}   & 4     & 8.00      & 68.45       & 4     & 8.00      & 74.24     \\
HAQ\cite{wang2019haq}      & $\backslash$    & $\backslash$ & $\backslash$     & MP    & 10.57     & 75.30     \\
BSQ\cite{yang2021bsq}      & $\backslash$    &  $\backslash$  &  $\backslash$  & MP    & 13.90     & 75.16     \\
\hline
\textbf{CSQ T2}    & \textbf{MP}  & \textbf{15.23}  & \textbf{69.11}    & \textbf{MP}   & \textbf{14.54}   & \textbf{75.25}     \\
\textbf{CSQ T3}    & \textbf{MP}    & \textbf{10.67}  & \textbf{69.73}    & \textbf{MP}   & \textbf{10.67}   & \textbf{75.47}     \\
\bottomrule
\end{tabular}
\endgroup
}
\end{table}

\subsection{Ablation study}
In this section, we discuss the key designs of the CSQ algorithm, including the comparison of the proposed continuous sparsification vs. STE in quantization-aware training (QAT), the effectiveness of the budget-aware model size regularization, and the control of accuracy-model size tradeoff. The quantization schemes obtained by CSQ under different model sizes are also demonstrated. All experiments in this section are conducted using ResNet-20 models~\cite{he2016deep} with 3-bit activation on the CIFAR-10 dataset~\cite{krizhevsky2009learning}.

{\noindent \textbf{Effectiveness of continuous sparsification.}} In this work, we replace the commonly used STE-based~\cite{bengio2013estimatingSTE} QAT with the proposed bit-level continuous sparsification in training quantized models. Table~\ref{table4} compares the QAT performance for a uniformly-quantized model trained with STE (STE-Uniform) and continuous sparsification (CSQ-Uniform). All models are trained from scratch with fixed-weight precision.
STE-Uniform follows the implementation in~\cite{polino2018model} where the floating-point latent weight is linearly quantized in the forward pass, and accumulates gradients in the backward pass with STE.
For CSQ-Uniform, we utilize the weight parameterization in Equation~(\ref{equ:bit}), where no bit mask is applied and only bit representations are trained. 
Under all precision, CSQ-Uniform outperforms STE-Uniform significantly, showing the effectiveness of bit-level continuous sparsification in leading to better convergence of quantized models. Additionally, as we propose to apply another level of continuous sparsification on the bit masks, the resulted bi-level continuous sparsification enables the proposed CSQ to find a better mixed-precision quantization scheme (CSQ-MP), which further boosts the performance over uniformly quantized counterparts.

\begin{table}[tb]
\small
\centering
\caption{CSQ vs. STE-based QAT performance. STE-Uniform training is implemented following~\cite{polino2018model}. }
\label{table4}
\begin{tabular}{c|cc} 
\toprule
\multicolumn{1}{l}{\textbf{W-Bits}}  & \textbf{QAT method} & \textbf{Accuracy (\%)}  \\
\hline
\multirow{3}{*}{4} 
& STE-Uniform~\cite{polino2018model}  & 88.89  \\
& CSQ-Uniform  & 91.93  \\
& \textbf{CSQ-MP}  & \textbf{92.68}  \\ 
\hline
\multirow{3}{*}{3}      
& STE-Uniform~\cite{polino2018model}   & 87.68  \\
& CSQ-Uniform    & 91.74  \\
& \textbf{CSQ-MP}    & \textbf{92.62}  \\ 
\hline
\multirow{3}{*}{\textbf{2}} 
& STE-Uniform~\cite{polino2018model}  & 84.35   \\
& CSQ-Uniform  &  91.67  \\
& \textbf{CSQ-MP}  & \textbf{92.34}   \\
\bottomrule
\end{tabular}
\vspace{-10pt}
\end{table}

\begin{figure}[tb]
\centering
\includegraphics[width=\linewidth]{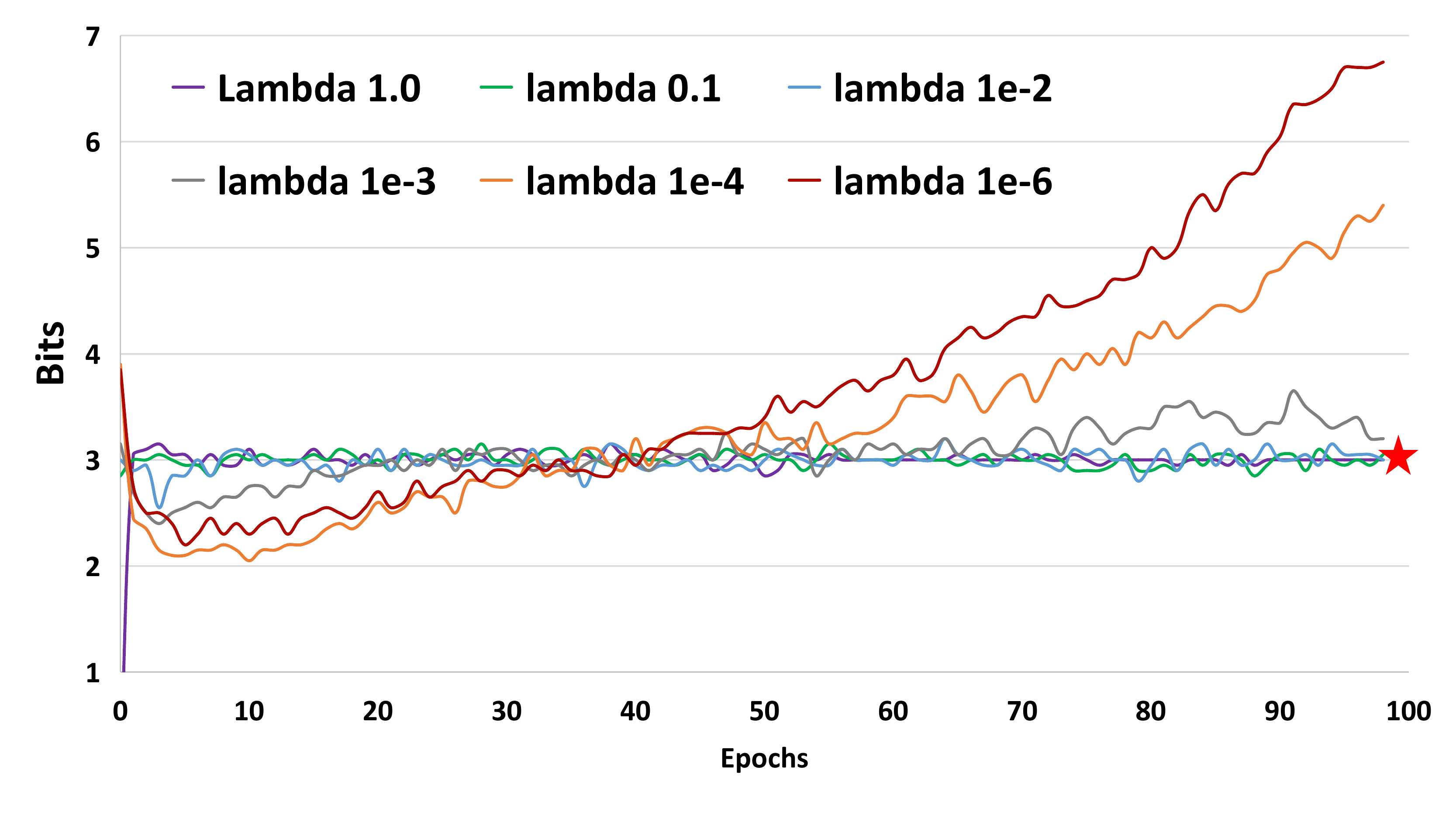}
\caption{Effect of base regularization strength $\lambda$ on the averaged model precision during training. All experiments are done with a target of 3-bit, as indicated by the ``red star''.}
\label{fig:lambda}
\vspace{-12pt}
\end{figure}

\begin{figure}[tb]
\centering
\includegraphics[width=\linewidth]{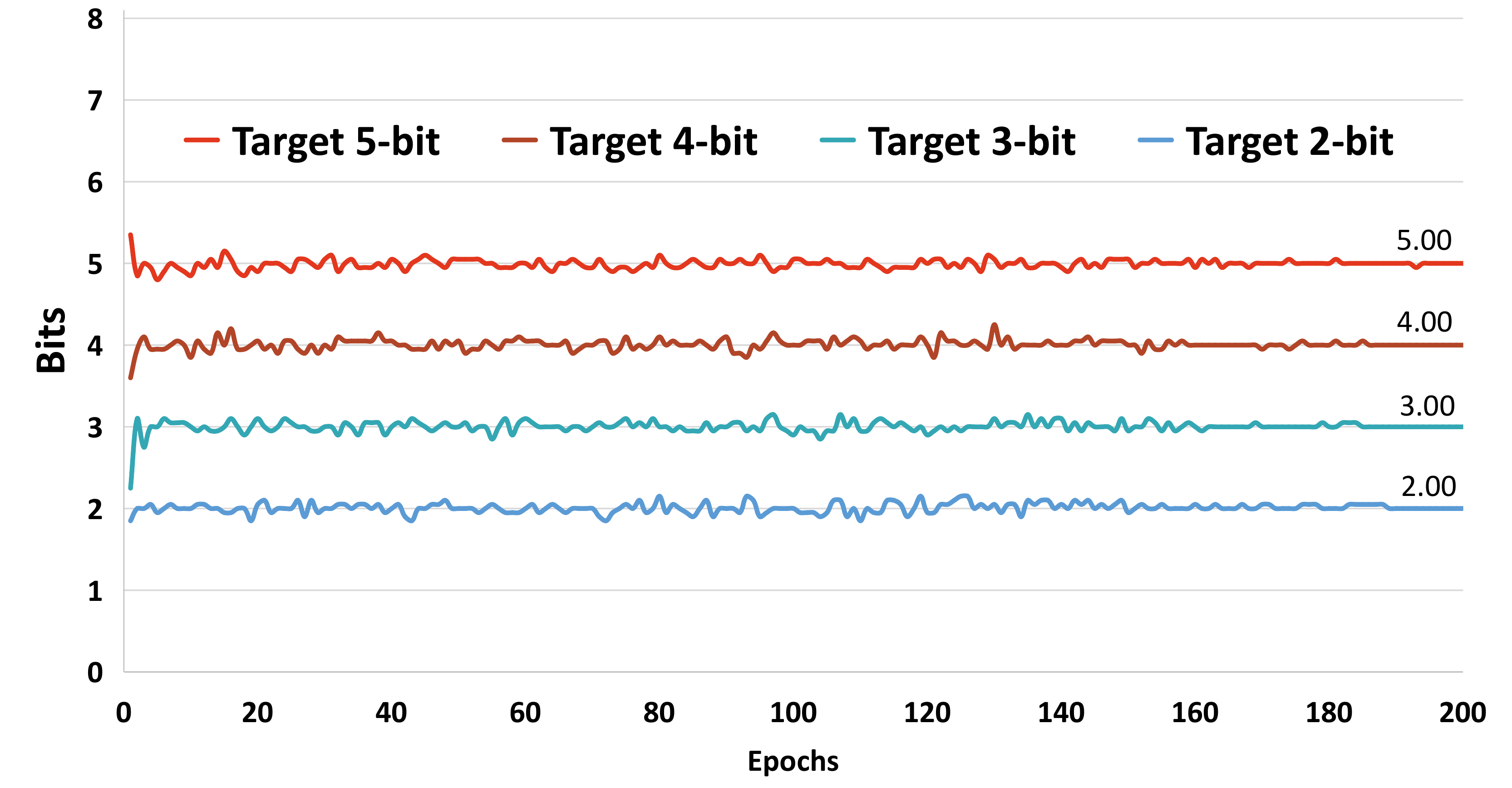}
\caption{Effect of different target precision on the averaged model precision during training.}
\label{fig:TargetTraining}
\vspace{-12pt}
\end{figure}

{\noindent \textbf{Budget-aware model size regularization.}}
As proposed in Section~\ref{ssec:Budget}, CSQ utilizes a budget-aware regularization to encourage the quantization scheme to meet a target precision. In this section, we explore the influence of the two hyperparameters in the regularization: base strength $\lambda$ and the target precision, on the final averaged precision achieved by CSQ.

Figure~\ref{fig:lambda} visualizes the changes in averaged precision of CSQ models trained with different initial values of $\lambda$. Note that since the bit masks are not exactly binary, we record the precision of each layer during training as $\sum_b\left[m_B^{(b)}\geq 0\right]$, as if the bit mask is gated with a step function. The target precision is set to be 3-bit for all trails. 
We note that the final model precision is not sensitive to the choice of $\lambda$ in a large range between 1e-3 and 1, where the model consistently converges to the desired target precision.  
$\lambda$ being too small (e.g. 1e-4 and 1e-6) resulted in less regularization strength to control the model precision effectively, which is as expected. To this end, we set $\lambda=0.01$ for all the experiments, which works well across different model architectures and datasets.  

Figure~\ref{fig:TargetTraining} shows the change of averaged model precision during CSQ training with different target precision. It can be seen that the proposed budget-aware regularization controls the model precision to be close to the target throughout the training process, and accurately converges to the target precision at the end of CSQ training.
The effectiveness of the regularization enables us to explicitly control the outcome of CSQ by setting an exact model size budget, mitigating the heuristic search of proper regularization strength for a specific model size required by previous methods~\cite{yang2021bsq,savarese2020winning}.
The stability of model precision throughout the training process also enables better convergence of the model, effectively leading to a mixed-precision model that is within the target budget while enjoying a preeminent accuracy.

{\noindent \textbf{Accuracy-model size trade-off.}} 
Performing CSQ training with different target precision effectively controls the size of the resulted quantization scheme, therefore exploring the tradeoff between model size and accuracy.
Table~\ref{table:tradeoff} summarizes the quantization results under different target bits obtained by CSQ using ResNet-20 on the CIFAR-10 dataset. Averaged precision (``Avg. prec.'') is computed across all model layers, and ``Comp($\times$)'' indicates the compression rate compared to the 32-bit floating-point model. The floating-point performance is also provided under the ``FP'' column for reference. 
As observed previously, the final average precision achieved by CSQ is fairly precise compared to the target. CSQ enables a lossless 5-bit quantization, while lower precision can be effectively achieved with the cost of a small accuracy drop.

\begin{table}[tb]
\small
\centering
\caption{Accuracy-size trade-off under different target bits.}
\label{table:tradeoff}
\begin{tabular}{c|ccccc|c} 
\toprule
Target          & 1-bit & 2-bit & 3-bit & 4-bit & 5-bit & FP \\ 
\hline
Ave. prec.   & 1.00   & 1.97  & 3.05  & 4.00  & 5.05 & 32  \\
Comp($\times$)  & 32    & 16.24 & 10.49 & 8.00  & 6.34 & 1  \\ 
\hline
CSQ acc.    & 90.33 & 91.70 & 92.42 & 92.51 & 92.61 & 92.62 \\
\bottomrule
\end{tabular}
\vspace{-10pt}
\end{table}

\begin{figure}[tb]
\centering
\includegraphics[width=\linewidth]{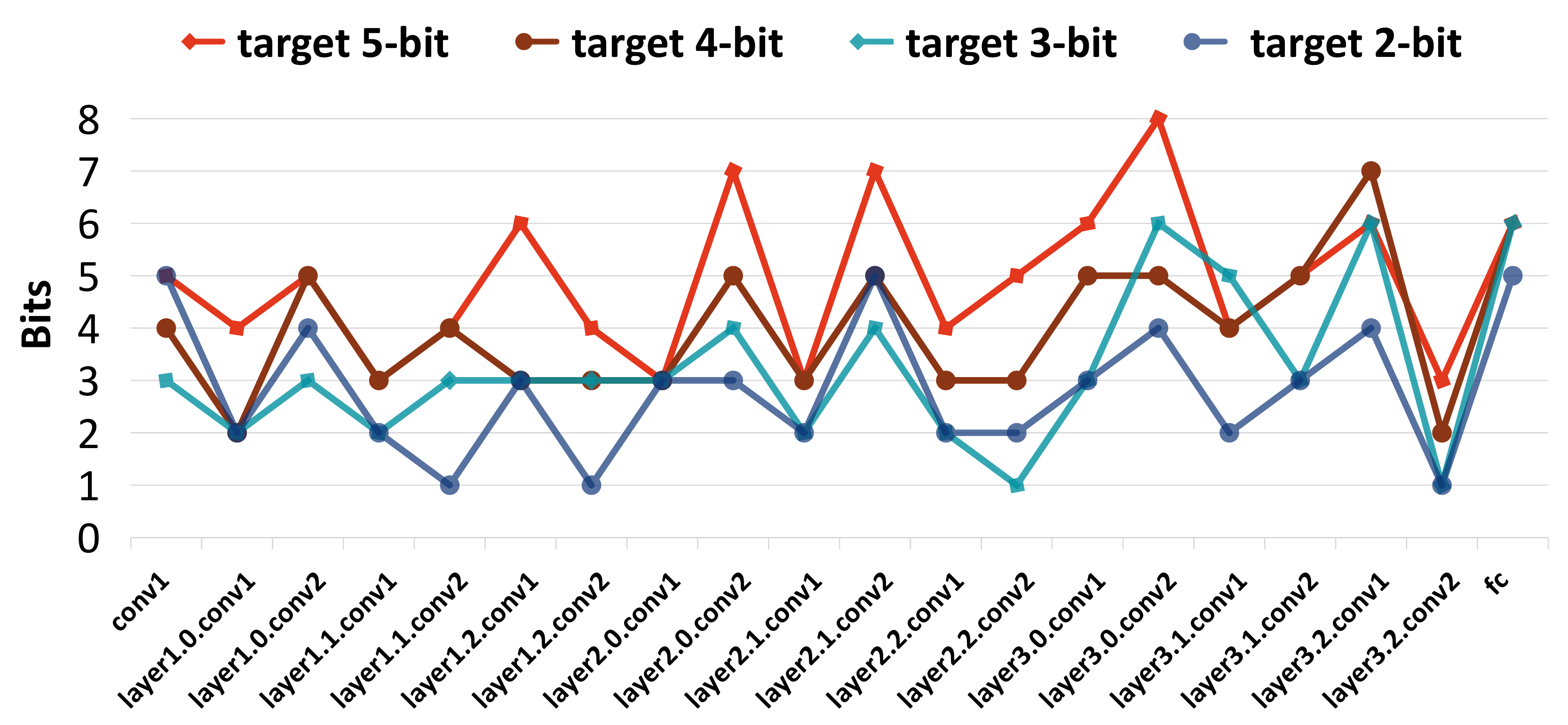}
\vspace{-12pt}
\caption{Layer-wise precision comparison of the quantization schemes produced by CSQ under different target bits;}
\label{fig:layersPrecision}
\vspace{-12pt}
\end{figure}

{\noindent \textbf{Layer-wise quantization schemes achieved with CSQ.}} Figure~\ref{fig:layersPrecision} shows the final precision of each layer in the mixed-precision quantization scheme obtained by CSQ under different target bits. Comparing among each other, the trends of quantization precision for the layers are generally consistent under different target bits, 
which echoes observations in previous mixed-precision quantization work~\cite{yang2021bsq,dong2019hawq}.
Interestingly, the importance ranking produced by CSQ is somewhat different from that of the previous work~\cite{yang2021bsq,dong2019hawq}. Our results show a roughly rising trend in precision from the input to the output layers, whereas the results from \cite{yang2021bsq,dong2019hawq} show a declining trend in precision, with the lowest precision in the final stage. Given the superior efficiency-accuracy tradeoff achieved by CSQ comparing to these methods, it shows that the heuristic-based importance criteria used in previous work may not accurately reflect the quantized model performance, while CSQ discovers better mixed-precision quantization schemes.

\section{Conclusion}

In this work, we propose CSQ, a novel method for bit-level mixed-precision DNN training using continuous sparsification. We improve the stability of training quantized DNNs with the bi-level continuous sparsification relaxation, and obtain mix-quantization schemes with explicit target precision utilizing the budget-aware model size regularization. 
Extensive experiments show the effectiveness of CSQ, where we achieve both higher accuracy and smaller quantization precision on various models and datasets comparing to state of the arts.

\bibliographystyle{unsrt}
\bibliography{main}
\end{document}